\documentclass[10pt, a4paper, dvipsnames]{article}

\usepackage{lrec-coling2024} 

\newcommand{\arxiv}[0]{\textsf{AI-arXiv}}
\newcommand{\gpt}[0]{\textsf{AI-GPT}}
\newcommand{\oa}[0]{\textsf{AI-OpenAlex}}
\usepackage{algorithm}
\usepackage{algorithmic}
\usepackage{colortbl}
\usepackage{comment}
\usepackage{tabularx}
\usepackage{booktabs}
\usepackage{array}

\usepackage{csquotes}

\definecolor{FaintGray}{gray}{0.925}
\name{Autumn Toney-Wails, Christian Schoeberl, James Dunham} 
\address{Georgetown University \\
\{autumn.toney, christian.schoeberl, james.dunham\}@georgetown.edu}
\begin{document}

\title{AI on AI: Exploring the Utility of GPT as an Expert Annotator of AI Publications}
\abstract{    Identifying scientific publications that are within a dynamic field of research often requires costly annotation by subject-matter experts. Resources like widely-accepted classification criteria or field taxonomies are unavailable for a domain like artificial intelligence (AI), which spans emerging topics and technologies. We address these challenges by inferring a functional definition of AI research from existing expert labels, and then evaluating state-of-the-art chatbot models on the task of expert data annotation. Using the arXiv publication database as ground-truth, we experiment with prompt engineering for GPT chatbot models to identify an alternative, automated expert annotation pipeline that assigns AI labels with 94\% accuracy. For comparison, we fine-tune SPECTER, a transformer language model pre-trained on scientific publications, that achieves 96\% accuracy (only 2\% higher than GPT) on classifying AI publications. Our results indicate that with effective prompt engineering, chatbots can be used as reliable data annotators even where subject-area expertise is required. To evaluate the utility of chatbot-annotated datasets on downstream classification tasks, we train a new classifier on GPT-labeled data and compare its performance to the arXiv-trained model. The classifier trained on GPT-labeled data outperforms the arXiv-trained model by nine percentage points, achieving 82\% accuracy. \\ \newline 
}
\maketitleabstract

\section{Introduction}
Analyzing scholarly literature provides insight into important features of a research field: identifying the active research community, tracking recent advances or breakthroughs, and mapping the translation of basic research into applications. A significant challenge for field-level analyses is the lack of clearly defined, widely accepted criteria to identify relevant scientific text, particularly when a field contains rapidly emerging topics and technologies \cite{kurzweil1985artificial,suominen2017exploring}. Artificial intelligence (AI) is one such research field. The challenge of identifying AI research is not new; in 1987 Schank wrote:

\begin{displayquote}
    Because of the massive, often quite unintelligible publicity that it gets, artificial intelligence is almost completely misunderstood by individuals outside the field. Even AI’s practitioners are somewhat confused about what AI really is \cite{schank1987ai}.
\end{displayquote}

Currently, almost 40 years after Schank questioned what AI was, identifying AI research is still an ambiguous task. Definitions of AI vary from academia, industry, and government, creating a challenge for researchers and policymakers when trying to conduct bibliometric studies, forecast technological capabilities, evaluate global leadership, or develop effective policy for AI systems and models \cite{kurzweil1985artificial,toney2022multi,grace2018will,krafft2020defining,cave2018ai}. While establishing a succinct definition or granular taxonomy of AI as a research field is outside the scope of this paper, we present an approach to automatically identify and classify AI research that leverages state-of-the-art large language models as expert annotators. We propose a generalizable framework for classification tasks that do not have a clearly defined labeling convention, and thus, are not amenable to costly, error-prone manual annotation. 

We derive an AI definition from published research activity over the past decade. Using a collection of author-identified AI research publications from the open-source arXiv database, we create a subset of AI-related scientific research publications, \arxiv{}, as a ground-truth labeled dataset. Our approach uses what subject-area experts have identified as relevant to AI research, reducing bias. The labels are assigned by the authors and thus reflect the evolution of the field; they are not bound by a static or dated definition.

With scientific publications' titles and abstracts as input text, we use two transformer language models pre-trained on scientific text, SciBERT \cite{Beltagy2019SciBERT} and SPECTER \cite{specter2020cohan}, for AI publication classification. For this classification task we use two scholarly literature datasets: 1) arXiv,\footnote{\url{https://arxiv.org}} as it contains author-assigned subject categories and 2) OpenAlex \cite{priem2022openalex}, as it contains the majority of scholarly literature but requires expert annotation. We establish an AI classification accuracy baseline using \arxiv{} to fine-tune both language models. Then, we explore the utility of OpenAI's GPT models as an automated expert annotator of AI publications using a zero-shot annotation prompt. We design a series of prompts with personas of varying levels of expertise (reader, researcher, and subject-matter expert) for the AI publication annotation task. We experiment with both GPT-3.5-Turbo and GPT-4 due to the significant cost difference.


We compare the GPT chatbot models' labeling accuracy and model performance to our baseline \arxiv{} classifier and evaluate how accurately GPT assigns labels to AI-related arXiv publications. Selecting the most reliable and accurate zero-shot prompt and GPT model, we generate a dataset of GPT-labeled publications, \gpt{}, to train a new publication classifier and compare it to the baseline \arxiv{} classifier. Our results show that both GPT-3.5-Turbo and GPT-4 achieve 94\% accuracy in data labeling, compared the baseline \arxiv{} classifier which achieves 96\% accuracy, suggesting that chatbots can be effectively used as expert annotators with reliable results. Evaluating the \arxiv{} and \gpt{} classifiers on a set of publications from the top 13 AI conferences, we find that the \gpt{} classifier outperforms the \arxiv{} classifier by nine percentage points, achieving 82\% accuracy.  

We summarize our contributions as follows: 1) we design an experimental framework to evaluate a chatbot's utility as a expert annotator, 2) we propose and discuss the merits of using a crowd-sourced AI definition derived from experts, and 3) we evaluate our framework on an AI research publication classification task.

\section{Background and Motivation}

Establishing a field definition or taxonomy is challenging for areas of research that encompass rapidly emerging topics and technologies, with AI being no exception. Since the term AI was first coined by McCarthy in 1955, researchers have addressed the question---what \textit{is} AI?---by surveying existing research and proposing frameworks and definitions \cite{schank1987ai,kurzweil1985artificial,russell2010artificial,fast2017long,martinez2018facets,krafft2020defining,shukla2019engineering}. We highlight two notable instances of computer scientists working to intentionally think through this question and answer it, both arriving at similar conclusions. Schank responded to this question pragmatically, stating that the definition of AI to a given researcher or organization depends directly on their research goals and methodology to design and implement their AI model. Kurzweil acknowledged that academia and industry are at odds with each other in forming a consensus about what AI is, correctly assessing that resolving this controversy is not likely in the near future. 

Although a clear, widely-accepted definition of AI has not been established, there are features of AI as a field that are agreed-upon, namely its direct relationship to machine learning (ML). Many publications have distinguished the two fields by describing ML as an application of AI \cite{alpaydin2016machine,groger2021there,woolridge}. In this work, we incorporate this notion, while we do not use the two terms (AI and ML) as synonyms, we consider ML to be a majority subset of AI research. Thus, we select publications with ML labels for our AI publication dataset and we include ML in our persona description for chatbot prompts.  

\subsection{Defining and Identifying AI Research}

Prior research has attempted to establish methodologies for identifying AI based on varying criteria. Goa et al. generate a list of 127 AI journals for publication analysis \cite{gao2021bibliometric}. In a similar approach, Martinez et al. study publications from AAAI and IJCAI conferences and Shukla et al. study publications from the Engineering Applications of Artificial Intelligence \cite{martinez2018facets,shukla2019engineering}. Using query-based methods, Niu et al. selected publications containing the term ``artificial intelligence'' and Miyazkai developed a query with 43 search terms (e.g., ``machine learning'' and ``facial recognition''). 

Scientific publication databases often provide research subject or topic areas that are not assigned manually by experts. One common example is the use of Microsoft Academic Graph's (MAG) fields of study, a hierarchy of research topics organized into five levels of granularity (level 0 - 4) \cite{sinha2015,shen-etal-2018-web}. These topic assignments allow for bibliometric analysis on research topics across all of science, but the topic assignment is unsupervised and based on embedding similarity. While this mitigates annotator bias, it relies on a static definition of AI derived from Wikipedia descriptions of concepts. Additionally, these assignments are not evaluated against ground-truth data.

Identifying AI research based on a set of publication venues or keywords restricts analysis to venues and terms that were relevant to the field at the time the study was conducted. These methods are also at risk of creating a narrowly scoped set of papers, ignoring more general or cross-disciplinary research that is relevant to the field. We address these shortcomings by selecting publications with author-assigned research categories, where the categories are assigned at the time of publication. This approach uses what subject-area experts consider to be relevant to AI and ML, incorporating how the field and its activity has evolved.

\subsection{Large Language Models as Expert Annotators}
With the ability to perform natural language processing tasks (e.g., document summarization, question and answering, and text classification) with human-like reasoning, LLMs' chatbots have enabled users to interact with artificially intelligent systems that can produce outputs indistinguishable from human-generated responses. However, despite chatbots' potential to respond with correct and reliable results, they are prone to respond with \textit{hallucinations}, a term representing seemingly random, non-factual or incoherent chatbot responses \cite{dziri-etal-2022-origin,ji2023survey}. OpenAI has continued to iterate on their GPT models to improve errors such as hallucinations, but their current state-of-the-art chatbot, GPT-4, is still susceptible to unreliable responses. \cite{openai2023gpt4}.

Chatbot hallucinations have resulted in research focused on how to effectively leverage a chatbot as a reliable data annotator through prompt engineering, as chatbots can be significantly less expensive than human annotators, but concerningly less reliable. Wang et al. find GPT-3 to be on average 10 times cheaper than human annotators when compared to Google Cloud Platform prices \cite{wang-etal-2021-want-reduce}. Gilardi et al. evaluated ChatGPT on over 6,000 tweets and news articles for numerous classification tasks, including relevance, topic assignment, and stance detection \cite{doi:10.1073/pnas.2305016120}. Using six trained annotators, they established ground-truth labels which they compared to Mechanical Turkers and ChaGPT and found that ChatGPT achieved higher annotation performance and was an estimated 30 times cheaper than manual annotators. 


Kim et al. evaluate ChatGPT's ability to label the strength of a claim as causal, conditional causal, correlational, or no relationship \cite{kim2023can}. The authors found that ChatGPT did not achieve state-of-the-art classification performance, concluding that chatbots have promising annotation capacity but improvement is needed for causal scientific reasoning. Our work analyzes chatbots' abilities when leveraged as a reliable, expert annotator on a zero-shot task. We consider our annotation task to be straight-forward, requiring less reasoning than evaluating causal scientific claims, but more expertise than a typical human annotator might have. Our annotation experiments investigate optimal prompt design for similar annotation tasks, presenting a format that can be adapted to other domains.

\section{Experimental Design}
\begin{figure}[ht!]
\centering
\includegraphics[width=\columnwidth]{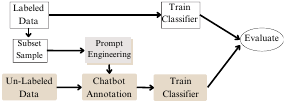}
\caption{Chatbot annotation experimental framework diagram.}
\label{fig:design}
\end{figure}

We design a framework to evaluate the utility of chatbots as expert annotators through prompt engineering and classification model performance, as shown in Figure \ref{fig:design}. Each step in this process can be adapted to a specific classification task, provided that there is a baseline labeled dataset or a comparable evaluation task for prompt engineering. In a small data task chatbot annotation can replace the classifier model. However, in this work we focus on tasks that use large datasets, which would be time-intensive and costly for a chatbot to annotate entirely.  

\subsection{Scientific Publication Classifier}
We experiment with two publicly available transformer language models: SciBERT and SPECTER. SciBERT is a pre-trained language model based on the Bidirectional Encoder Representations from Transformers (BERT) model and trained on a sample of 1.14M papers from Semantic Scholar (using full-text) \cite{Beltagy2019SciBERT}. The Scientific Paper Embeddings using Citation-informed TransformERs (SPECTER) improves on SciBERT by incorporating the citation graph that exists between academic publications \cite{specter2020cohan}. Compared to SciBERT, SPECTER decreases training times while maintaining (or exceeding) performance, particularly for classification tasks.

We implement all scientific publication classifiers in the same way, changing transformer models and datasets for each experiment. All non-default parameters are specified in the Supplementary Materials\footnote{All code and data will be available on GitHub upon publication}. We use train, test, and validation dataset splits of 70\%, 15\%, and 15\% respectively. 

\subsection{Data Annotation Prompt Engineering}
To automatically identify AI research publications we leverage state-of-the-art LLMs' chatbot feature to assign binary (AI or Non-AI) labels given a publication's title and abstract. This requires experimentation in prompt engineering to select a prompt that will produce reliable labels that are usable on their own (e.g., treating the chatbot as the classifier) or functional for generating training data (e.g., creating a labeled dataset). We aim to identify a single, zero-shot prompt that will produce accurate and parsable responses to initiate an automated annotation pipeline. 

We use the GPT-3.5-Turbo and GPT-4 chatbots, as GPT-4 is a more robust model but is approximately 20 times more expensive to query than GPT-3.5-Turbo. Using the \texttt{openai} Python package, we run our prompt engineering experiments with \texttt{temperature} set to 0, which indicates that the most likely output from the model should be selected. We design a series of prompts that provide increasing specificity of the AI expertise we ask the GPT chatbot models to personify: a reader, a researcher, and a subject-matter expert. Table \ref{tab:prompts} gives the nine prompts that we test. Each of the three personas are tested with variations to the instructions, providing the chatbot with awareness of non-relevant publications and clarity on how to annotate. These prompts result in consistent and parsable responses, enabling us to compare their performance across chatbot models and prompts. 

\newcolumntype{R}{>{\raggedright\arraybackslash}p{1in}}
\begin{table*}[ht!]
    \small
    \centering
    \begin{tabularx}{\textwidth}{RX}
         \toprule
         \textbf{Prompt Type} & \textbf{Prompt}\\ \midrule
        Reader, Researcher, Expert & You are a \textbf{[persona type]} in AI/ML, and you are given an annotation task. Given a publication's title and abstract, assign a AI or Non-AI label determining if the publication belongs to the field of AI/ML research and a predicted probability of relevance. Provide just the label and prediction in your answer. 
        
        \\
        
        
        Uncertainty & You are a \textbf{[persona type]} in AI/ML, and you are given an annotation task. Given a publication's title and abstract, assign a AI or Non-AI label determining if the publication belongs to the field of AI/ML research and a predicted probability of relevance. \textbf{Some papers may be in STEM fields but not exactly AI, please assign AI only if you are confident.} Provide just the label and probability in your answer. 
        
        \\
        Uncertainty and clarity &  You are a \textbf{[persona type]} in AI/ML, and you are given an annotation task. Based on the title and abstract of an academic publication, assign a label ‘AI’ or ‘Non-AI’ indicating whether the publication belongs to the field of AI/ML research. Also assign a score between 0 and 1 that describes how confident you are in the label. \textbf{Some papers may be in STEM fields but not exactly in AI/ML. Please assign the ‘AI’ label only if you are confident. Otherwise, assign the ‘Non-AI’ label and quantify your uncertainty in the score}. Respond only with the label and the score.\\

         \bottomrule
    \end{tabularx}
\caption{Zero-shot chatbot prompt variations for reliable data annotation experiments.}
\label{tab:prompts}
\end{table*}


\subsection{Classifier Performance Evaluation}
We include an evaluation task that is separate from comparing the validation performance of the two classification models. This evaluation task should compute model performance on a new dataset to evaluate the generalizability of the models, as one goal of using chatbots for data annotation is generating a more representative training dataset. For our domain application, we follow \citet{toney2022multi} and use a set of publications that appeared in one of the 13 top AI and ML conferences from CSRankings\footnote{\url{csrankings.org}}: AAAI Conference on Artificial Intelligence (AAAI), International Joint Conference on Artificial Intelligence (IJCAI), IEEE Conference on Computer Vision and Pattern Recognition (CVPR), European Conference on Computer Vision (ECCV), IEEE International Conference on Computer Vision (ICCV), International Conference on Machine Learning (ICML), International Conference on Knowledge Discovery and Data Mining (SIGKDD), Neural Information Processing Systems (NeurIPS), Annual Meeting of the Association for Computational Linguistics (ACL), North American Chapter of the Association for Computational Linguistics (NAACL), Conference on Empirical Methods in Natural Language Processing (EMNLP), International Conference on Research and Development in Information Retrieval (SIGIR), International Conference on World Wide Web (WWW).

\section{Scholarly Literature Datasets}
We use two open-source scholarly literature datasets in our experiments: arXiv and OpenAlex. Here we describe the details of each data source, as well as define how we generate our three AI publication datasets for experimentation: \arxiv{}, \gpt{}, and \oa{}.

\paragraph{arXiv Dataset}
Hosting over 2 million scientific publications across 8 research fields (computer science, economics, electrical engineering and systems science, mathematics, physics, quantitative biology, quantitative finance, and statistics), arXiv is a useful resource for classification tasks on scientific text, as all publications are assigned research area categories by authors. arXiv's Computing Research Repository (CoRR) lists 39 sub-categories including artificial intelligence and machine learning. CoRR editors review each publication and its assigned research categories, so we treat these labels as expert-annotated.

\begin{figure}[ht!]
\centering
\includegraphics[width=\columnwidth]{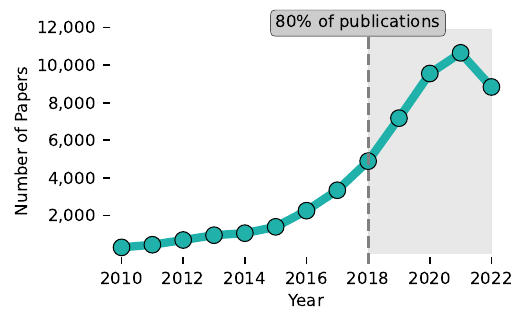}
\caption{Number of AI arXiv by publication year. Data accessed on 10-13-2022, thus 2022 is incomplete.}
\label{fig:arxiv_overtime}
\end{figure}

We generate the \arxiv{} dataset, comprising of publications since 2010 that CoRR identifies as being AI-related using their author-assigned research categories.\footnote{\url{https://arxiv.org/category_taxonomy}} We include a publication in the \arxiv{} dataset if it was labeled with at least one of the following research topics: Artificial Intelligence (\texttt{cs.AI}), Computation
and Language (\texttt{cs.CL}), Computer Vision and Pattern Recognition (\texttt{cs.CV}), Machine Learning (\texttt{cs.LG, stat.ML}), Multiagent Systems (\texttt{cs.MA}), and Robotics (\texttt{cs.RO}). The dataset begins in 2010, but the majority of AI-related publications (80\%) are from 2018 and later, as shown in Figure \ref{fig:arxiv_overtime}. This graph illustrates the rapid influx of AI-related publications in the past five years, but also highlights that while the \arxiv{} contains publications over the past decade it mainly represents research activity from the past five years.

Authors can assign a primary subject category and cross-post under additional categories (i.e., a publication may be assigned more than one category). Table \ref{tab:arxiv_categories} displays the number of publications by their primary arXiv research category and by cross-post research categories. The top two most frequent categories are machine learning (\texttt{cs.LG}) with 120,525 publications and computer vision (\texttt{cs.CV}) with 82,760 publications. Mutliagent systems (\texttt{cs.MA}) is the least frequent category with 5,304 publications. 

{
\newcolumntype{a}{>{\columncolor{FaintGray}}r}

\begin{table}[ht]
    \small
    \centering
    \setlength{\tabcolsep}{1.0pt}
    \begin{tabularx}{\columnwidth}{@{}Xrararar}
    \toprule
    {\bf Type} & {\tt AI} & {\tt CL} & {\tt CV} & {\tt LG} & {\tt MA} & {\tt RO} & {\tt ML} \\ \midrule
    P & 14,615 & 28,914 & 63,016 & 55,991 & 1,801 & 14,246 & 13,516 \\
    CP & 36,332 & 7,860 & 19,744 & 64,534 & 3,503 & 6,381 & 42,412 \\
    \bottomrule
    \end{tabularx}
    
    \caption{Number of publications by primary (P) and cross-post (CP) research category type.}
    \label{tab:arxiv_categories}
\end{table}
}

\paragraph{OpenAlex Dataset}
Containing over 240 million scientific publications across all fields of science, we use OpenAlex as our un-labeled publication dataset that we sample from for our chatbot annotation task. To maintain consistency with the \arxiv{} dataset, we restrict publication year to 2010 or later. To refine the OpenAlex publication set that we sample from, we require at least one citation per publication, resulting in 114,635,253
publications. We create the \gpt{} dataset using the 76K sampled OpenAlex publications that were assigned AI labels by the GPT models. 

OpenAlex also provides concepts\footnote{\url{https://docs.openalex.org/api-entities/concepts}} associated with each publication which are structured similarly to Microsoft Academic Graph's field of study taxonomy. These concepts are automatically assigned to publications by document embedding similarity and are not validated against ground truth data for all possible field categories, therefore the assignments contain noise. There are 19 concepts at the most general level (e.g., mathematics and computer science) and there are 283 subtopics at the next level of granularity (e.g., artificial intelligence and machine learning). Because these concepts are commonly used for labels in classification tasks---and specifically used by \citet{specter2020cohan} to evaluate SPECTER---we generate a set of AI publications to compare against the \arxiv{} dataset. We select concepts that directly map to the arXiv categories; publications with artificial intelligence, computer vision, machine learning, and natural language processing listed as the top field compose the \oa{} dataset, totaling 4,305 AI publications.

\section{Results and Evaluation}

\subsection{\arxiv{} Classifier}
Using the \arxiv{} dataset we train and evaluate the SciBERT and SPECTER language models on a binary classification task (AI or non-AI) given a publication's title and abstract. We run this initial experiment on a 10\% sample of the \arxiv{} dataset (over 150K publications) fine-tuning both SciBERT and SPECTER, and on the \oa{} dataset fine-tuning SPECTER. We only fine-tune SciBERT on the \arxiv{} dataset, as Cohan et al. showed that SPECTER (achieving 80\%) outperformed SciBERT (achieving 72\%) on classifying publications by MAG's most general fields of study. Table \ref{tab:arxiv_model_performance} displays the three classifier performance metrics.

\begin{table}[ht]
    \small
    \centering
    \setlength{\tabcolsep}{4pt}
    \begin{tabularx}{\columnwidth}{Xcccc}
         \toprule
         \textbf{Model} & \textbf{Accuracy} & \textbf{Precision} & \textbf{Recall} & \textbf{F1} \\ \midrule
        SciBERT & 0.96 & 0.88 & 0.86 & 0.87 \\
        SPECTER$_{MAG}$ & 0.93 & 0.93 & 0.60 & 0.73 \\
        SPECTER$_{arXiv}$ & 0.96 & 0.87 & 0.88 & 0.88\\
         \bottomrule
    \end{tabularx}
\caption{10\% sample the \arxiv{} dataset and the \oa{} dataset model performances with SciBERT and SPECTER.}
\label{tab:arxiv_model_performance}
\end{table}

Table \ref{tab:arxiv_model_performance} shows that SPECTER fine-tuned on \arxiv{} outperforms the other two models in all performance metrics, with approximately the same training time and cost as SciBERT, thus we select SPECTER as our transformer language model for classification tasks. Table \ref{tab:arxiv_model_splits} displays the total counts for the train, test, and validation splits for the full \arxiv{} dateset. The full \arxiv{} classifier produces similar results to the 10\% sample model in Table \ref{tab:arxiv_model_performance}: 0.89 precision, 0.87 recall, and 0.88 F1.



\begin{table}[ht]
    \centering
    \small
    \setlength{\tabcolsep}{12pt}
    \begin{tabularx}{\columnwidth}{Xrrr}
         \toprule
         \textbf{Split} & \textbf{Total Count} &\textbf{AI} & \textbf{Non-AI}  \\ \midrule
        Train & 1,093,647 & 173,292 & 920,355  \\
        Test & 234,353 & 37,136 & 197,217 \\
        Validation & 234,353 & 37,042 & 197,311 \\
        
         \bottomrule
    \end{tabularx}
\caption{Train, test, and validation dataset split counts for the \arxiv{} dataset by labels.}
\label{tab:arxiv_model_splits}
\end{table}

\subsection{GPT for Data Annotation}
\begin{figure*}[ht!]
\centering
\includegraphics[width=\textwidth]{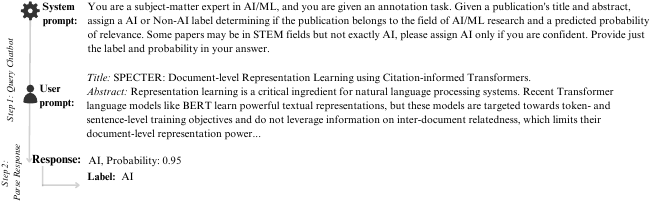}
\caption{GPT annotation example prompts and response.}
\label{fig:gpt_example}
\end{figure*}

With the nine prompt variations listed in Table \ref{tab:prompts}, we experiment with GPT-3.5-Turbo and GPT-4 as expert data annotators. We sample 5,000 publications from the \arxiv{} dataset, and due to cost we only query GPT-4 with 2,500 publications for our initial chatbot comparison. Figure \ref{fig:gpt_example} illustrates the annotation pipeline using the expert with uncertainty and clarity system prompt as an example. In all chatbot responses, we are able to parse the response into a binary label (AI/Non-AI) and a predicted relevance probability (with the values always ranging between 0 and 1).

The system prompt significantly impacts a chatbot's ability to accurately annotation publications, as shown in Table \ref{tab:gpt_annotation_results}. GPT-4 provides consistent performance across all prompt variations; however, the best GPT-3.5-Turbo prompts are able to produce the same accuracy results. We find that GPT-3.5-Turbo has a stronger improvement when including language surrounding uncertainty and clarity in the prompt. Adding additional prompt language regarding the annotation task increases accuracy by 13.8 percentage points on average; however, the persona shifts have minimal effect across both chatbot models and prompt variation. In contrast to GPT-3.5-Turbo, GPT-4 has significantly less improvement when including the uncertainty and clarity clauses. 

{
\newcolumntype{a}{>{\columncolor{FaintGray}}c}
\begin{table}[ht]
    \small
    \centering
    \setlength{\tabcolsep}{3.3pt}
    \begin{tabularx}{\columnwidth}{Xcccaaaccc}
    \toprule
    {\bf GPT Model} & \multicolumn{9}{c}{\bf Accuracy} \\ \cmidrule(lr){2-10}
    & \multicolumn{3}{c}{Reader} & \multicolumn{3}{c}{Researcher} & \multicolumn{3}{c}{Expert} \\ \cmidrule(lr){2-4} \cmidrule(lr){5-7} \cmidrule(lr){8-10}
    & - & +U & +UC & - & +U & +UC & - & +U & +UC \\ \midrule
    3.5-T & .79 & .91 & .92 & .76 & .91 & .92 & .78 & .91 & .92 \\
    4 & .91 & .94 & .92 & .91 & .92 & .92 & .91 & .94 & .94 \\
    \bottomrule
    \end{tabularx}
\caption{GPT chatbot model comparison across all nine prompts for data annotation. +U denotes the prompt including uncertainty  and +UC denotes the prompt including uncertainty and clarity.}
\label{tab:gpt_annotation_results}
\end{table}
}

We select the expert with uncertainty and clarity prompt as our best prompt for further experimentation. Investigating if there was a subject area difference between GPT-3.5-Turbo's and GPT-4's performance, we compared the annotation accuracy of the best prompt. Table \ref{tab:gpt_arxiv_accuracy} displays the accuracies by arXiv research category across both GPT models. GPT-3.5-Turbo has the highest annotation accuracies on machine learning (89\%), NLP (89\%), and computer vision (87\%) with multiagent systems (67\%) and robotics (75\%) categories having the worst performance. Similarly, GPT-4 performs the worst on multiagent systems (89\%) and robotics (84\%), but performs the best on machine learning (99\%) and AI (98\%).


\begin{table}[t]
    \small
    \centering
    \setlength{\tabcolsep}{3pt}
    \begin{tabularx}{\columnwidth}{Xccccccccc}
    \toprule
    {\bf GPT Model} & \multicolumn{9}{c}{\bf Accuracy} \\ \cmidrule(lr){2-10}
     & {\tt AI} & {\tt CL} & {\tt CV} & {\tt LG} & {\tt MA} & {\tt RO} & {\tt ML} & None & Overall \\ \midrule
    3.5-T & .84 & .89 & .87 & .89 & .67 & .75 & .81 & .93 & .92 \\
    4 & .98 & .93 & .93 & .99 & .89 & .84 & .97 & .94 & .94 \\ \bottomrule
    \end{tabularx}
\caption{Labeling accuracies by GPT model and arXiv categories. This represents primary and cross-posted categories, thus publications can be counted in multiple categories.}
\label{tab:gpt_arxiv_accuracy}
\end{table}

\begin{figure}[ht!]
\centering
\includegraphics[width=\columnwidth]{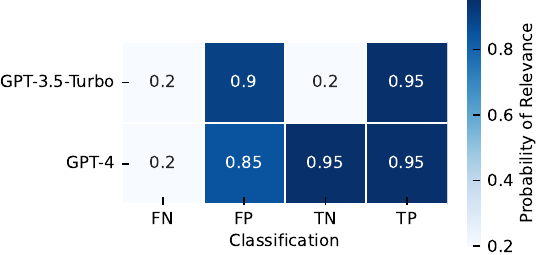}
\caption{Median predicted probability of relevance by GPT model across classification types.}
\label{fig:probs}
\end{figure}

We evaluate the predicted probabilities of relevance that the chabots provided. All responses included a probability value, with all values being correctly bounded by 0 and 1. The majority of values were either 0.95 or 0.2, thus we present the median predicted probably by GPT model and predicted class in Figure \ref{fig:probs}. We expect low median values for false negatives (FN) and true negatives (TN), indicating that the chatbot interpreted the publication's title and abstract as being non-relevant to AI research. We find that GPT-3.5-Turbo has expected results, with the negative class having low predicted probability (0.2 for FN and TN) and the positive class having high predicted probabilities (0.95 for TP and 0.9 for FP). In contrast, GPT-4 has the highest predicted probabilities for true positive (0.95), true negative (0.95), and false positive (0.85), with only a low predicted probability for false negatives (0.2). 

\subsection{\gpt{} Classifier}
We implement the same SPECTER model used for the \arxiv{} classifier, but substitute in the \gpt{} dataset. Table \ref{tab:gpt_model_splits} displays the train, test, and validation \gpt{} dataset splits, with a significant class imbalance towards non-AI papers. However, this is consistent with the representation of peer-reviewed AI publications in the context of all of science, with AI publications representing approximately 3.8\% of all scientific literature \cite{zhang2021ai}. The gpt{} classifier achieves 92\% overall accuracy with 0.70 precision, 0.92 recall, and 0.80 F1.

\begin{table}[ht]
    \centering
    \small
    \setlength{\tabcolsep}{12pt}
    \begin{tabularx}{\columnwidth}{Xrrr}
         \toprule
         \textbf{Split} & \textbf{Total Count} & \textbf{AI} & \textbf{Non-AI}  \\ \midrule
        Train & 53,459 & 1,288 & 52,171\\
        Test & 11,456 & 276 & 11,180\\
        Validation & 11,455 & 276 & 11,179 \\
        
         \bottomrule
    \end{tabularx}
\caption{Train, test, and validation dataset split counts for the \gpt{} dataset by labels.}
\label{tab:gpt_model_splits}
\end{table}

\subsection{Classifier Evaluation}
In order to evaluate the utility of chatbots as expert annotators and AI arXiv publications as a functional definition of AI research, we compare the \gpt{}, \arxiv{}, and \oa{} classifiers on a new dataset that contains publications from the top 13 AI conferences. Table \ref{tab:model_eval} presents the models' accuracies by conference venue as well as the overall accuracy considering all AI conference papers as a set. We find that the \gpt{} model outperforms the \oa{} and \arxiv{} model across all venues, producing an overall accuracy of 82\%. While accuracy by conference varies significantly across the classifiers, we find that CVPR, EMNLP, and ICCV have the highest accuracies for all three classifiers and that WWW and SIGIR have the lowest. None of the classifiers perform the best on AAAI or IJCAI, which are the two most explicitly AI-related conferences. 

\newcolumntype{S}{>{\raggedright\arraybackslash}m{0.75in}}
\newcolumntype{M}{>{\raggedleft\arraybackslash}m{0.47in}}
\newcolumntype{K}{>{\raggedleft\arraybackslash}m{0.42in}}

\newcolumntype{N}{>{\raggedleft\arraybackslash}m{0.75in}}

\newcolumntype{T}{>{\raggedleft\arraybackslash}m{0.5in}}

\begin{table}[ht]
    \small
    \setlength{\tabcolsep}{4pt}
    \centering
    \begin{tabularx}{\columnwidth}{XTNMK}
         \toprule
         \textbf{Venue} & \textbf{Num. Papers} & \textbf{\oa{} Acc.}& \textbf{\arxiv{}  Acc.} & \textbf{\gpt{} Acc.}\\ \midrule
        NeurIPS & 10,999 & 0.45 & 0.73 & 0.84\\
        AAAI & 10,446 & 0.53 & 0.88 & 0.89 \\
        IJCAI & 9,700 & 0.32 & 0.68 & 0.70\\
        ICML & 6,192 & 0.46 & 0.70 & 0.82 \\
        CVPR & 3,381 & 0.89 & 0.86 & 0.95 \\
        SIGIR & 2,492 & 0.16 & 0.21 & 0.59\\
        NAACL & 1,834 & 0.62 & 0.86 & 0.93 \\
        ICCV & 1,619 & 0.92 & 0.90 & 0.99 \\
         WWW & 1,241 & 0.12 & 0.21 & 0.45 \\
         SIGKDD & 1,064 & 0.31 & 0.60 & 0.84 \\
         ACL & 932 & 0.71 & 0.86 & 0.93 \\
         EMNLP & 839 & 0.79 & 0.93 & 0.95 \\
         ECCV & 64 & 0.67 & 0.28 & 0.72 \\
         \midrule
         Overall & & 0.48 & 0.73 & 0.82 \\
         \bottomrule
    \end{tabularx}
\caption{\oa{}, \arxiv{}, and \gpt{} classification accuracies by conference venue.}
\label{tab:model_eval}
\end{table}

\section{Discussion}

\paragraph{Defining Research Fields via Expert Crowd-sourcing: }

A known limitation in text classification for AI research is the key step of identifying a set of labeled publications for classifier training. Designing a manual annotation task with either few expert or many non-expert annotators places the responsibility of defining what AI is on the authors, as they need to develop annotation instructions for the labeling task. Implementing unsupervised natural language processing techniques, such as topic modeling or document embedding clustering, lacks the transparency and reproducibility that can be achieved with a supervised classification model using reliably labeled data. 

Our approach treats the authors as experts in the field and considers the publications and potential evolution of how authors assign labels as a time-relevant representation of research activity, opposed to relying on a static or narrow field definition. With our \arxiv{} classifier results, we find that our ground-truth data is functional as training data for a generalized AI classification model in comparison to the \oa{} dataset, which labels publications by document embedding similarity. The \arxiv{} model is restricted to negative (non-AI) samples that are in STEM fields, meaning that the separation between relevant and non-relevant AI research is scoped to publications that would be more likely to blur the field's boundary lines. In comparison, the \oa{} and \gpt{} datasets have negative samples that span all of science, which could prove more optimal for classifier generalizability.

\paragraph{Evaluating the Utility of Chatbots as Annotators:}

A notable challenge when working with chatbots is their inconsistency and tendency to hallucinate responses. These two flaws are of particular issue in an annotation task, where reliable responses and consistent reasoning are necessary. An additional challenge when working with chatbots, is the lack of transparency in the LLM’s training data. For example, some tasks might prove to be more suitable to chatbots as annotators because the underlying LLM was trained on large amounts of relevant data, or in our annotation task, the same data we are asking GPT to label. 

While we do not explore in-depth methods to try to uncover what scientific publications GPT models might have observed during training, we do provide various ways of analyzing the responses from GPT. Our first approach is experimenting with variations in prompts with personas of increasing expertise. These experiments aim to discover what type of scoping chatbots need in system prompts in order to produce reliable responses. Next, we look at the annotation accuracy by subfields of AI to understand if there are any particular subsets of the data that GPT struggles with. These results show that GPT-4 achieves exceptionally high accuracy with open-source data labeled explicitly as AI, leading to speculation that this accuracy might be due to data memorization rather than model reasoning.
 
Our annotation results from prompt engineering indicate that while selecting the right persona is important for GPT, it is also necessary to design a prompt that encourages reasoning. We found that including specific uncertainty and clarity clauses in our prompts boosted GPT-3.5-Turbo's performance to be comparable with GPT-4, whereas the changes in expertise did not. Additionally, we explored including instructions for the chatbot to respond with a predicted probability of relevance for every title and abstract. While both GPT models consistently understood the task of responding with a probability, and always with a value between 0 and 1, GPT-3.5 responded more reliably. We found that GPT-4 responded with a median probability of 0.95 for labeling non-AI publications correctly, indicating that either GPT-4 did not comprehend the prompt instructions (i.e., the probability is representing annotation confidence) or that its responses are not associated with an internalized probability.

\section{Conclusion}

In this work we investigate the utility of a chatbots as expert annotators by evaluating their annotation agreement with ground-truth data and their model performance on downstream classification tasks for identifying AI research publications. We address the challenge of identifying AI, a rapidly emerging research field with no clear definition, by leveraging expert, crowd-sourced data on arXiv. We find that GPT models are able to achieve high accuracy as expert annotators on AI publications, producing reliable and parsable responses necessary for an annotation task. Our prompt engineering experiments indicate that chatbots have the highest performance when the prompt includes a relevantly-scoped persona (e.g., AI researcher or subject-matter expert) as well as details on how to consider edge cases (e.g., language describing how to consider uncertainty or providing clarity on the annotation task). We also find that GPT labels can be reliably used in downstream classification tasks as training data. Our experiments showed that even in large datasets with no underlying labels, GPT can provide a functional boundary between positive and negative examples. Collectively, these findings signal the ability of LLMs to provide scalable and efficient data annotation for bibliometric analysis, upon which more complex models can be built. 


\nocite{*}
\section{Bibliographical References}\label{sec:reference}

\bibliographystyle{lrec-coling2024-natbib}
\bibliography{biblio}

\begin{thebibliography}{37}
\expandafter\ifx\csname natexlab\endcsname\relax\def\natexlab#1{#1}\fi

\bibitem[{goo()}]{google}

\newblock \href
  {https://cloud.google.com/learn/artificial-intelligence-vs-machine-learning}
  {Artificial intelligence (ai) vs. machine learning (ml)}.

\bibitem[{Alpaydin(2016)}]{alpaydin2016machine}
Ethem Alpaydin. 2016.
\newblock \emph{Machine learning: the new AI}.
\newblock MIT press.

\bibitem[{Beltagy et~al.(2019)Beltagy, Lo, and Cohan}]{Beltagy2019SciBERT}
Iz~Beltagy, Kyle Lo, and Arman Cohan. 2019.
\newblock \href {http://arxiv.org/abs/arXiv:1903.10676} {Scibert: Pretrained
  language model for scientific text}.
\newblock In \emph{EMNLP}.

\bibitem[{Cave and {\'O}h{\'E}igeartaigh(2018)}]{cave2018ai}
Stephen Cave and Se{\'a}n~S {\'O}h{\'E}igeartaigh. 2018.
\newblock An ai race for strategic advantage: rhetoric and risks.
\newblock In \emph{Proceedings of the 2018 AAAI/ACM Conference on AI, Ethics,
  and Society}, pages 36--40.

\bibitem[{Clement et~al.(2019)Clement, Bierbaum, O'Keeffe, and
  Alemi}]{clement2019use}
Colin~B Clement, Matthew Bierbaum, Kevin~P O'Keeffe, and Alexander~A Alemi.
  2019.
\newblock On the use of arxiv as a dataset.
\newblock \emph{arXiv preprint arXiv:1905.00075}.

\bibitem[{Cohan et~al.(2020)Cohan, Feldman, Beltagy, Downey, and
  Weld}]{specter2020cohan}
Arman Cohan, Sergey Feldman, Iz~Beltagy, Doug Downey, and Daniel~S. Weld. 2020.
\newblock {SPECTER: Document-level Representation Learning using
  Citation-informed Transformers}.
\newblock In \emph{ACL}.

\bibitem[{Dziri et~al.(2022)Dziri, Milton, Yu, Zaiane, and
  Reddy}]{dziri-etal-2022-origin}
Nouha Dziri, Sivan Milton, Mo~Yu, Osmar Zaiane, and Siva Reddy. 2022.
\newblock \href {https://doi.org/10.18653/v1/2022.naacl-main.387} {On the
  origin of hallucinations in conversational models: Is it the datasets or the
  models?}
\newblock In \emph{Proceedings of the 2022 Conference of the North American
  Chapter of the Association for Computational Linguistics: Human Language
  Technologies}, pages 5271--5285, Seattle, United States. Association for
  Computational Linguistics.

\bibitem[{Fast and Horvitz(2017)}]{fast2017long}
Ethan Fast and Eric Horvitz. 2017.
\newblock Long-term trends in the public perception of artificial intelligence.
\newblock In \emph{Proceedings of the AAAI conference on artificial
  intelligence}, volume~31.

\bibitem[{Gao et~al.(2021)Gao, Jia, Zhao, Chen, Xu, Geng, and
  Song}]{gao2021bibliometric}
Fang Gao, Xiaofeng Jia, Zhiyun Zhao, Chih-Cheng Chen, Feng Xu, Zhe Geng, and
  Xiaotong Song. 2021.
\newblock Bibliometric analysis on tendency and topics of artificial
  intelligence over last decade.
\newblock \emph{Microsystem Technologies}, 27:1545--1557.

\bibitem[{Gilardi et~al.(2023)Gilardi, Alizadeh, and
  Kubli}]{doi:10.1073/pnas.2305016120}
Fabrizio Gilardi, Meysam Alizadeh, and Maël Kubli. 2023.
\newblock \href {https://doi.org/10.1073/pnas.2305016120} {Chatgpt outperforms
  crowd workers for text-annotation tasks}.
\newblock \emph{Proceedings of the National Academy of Sciences},
  120(30):e2305016120.

\bibitem[{Grace et~al.(2018)Grace, Salvatier, Dafoe, Zhang, and
  Evans}]{grace2018will}
Katja Grace, John Salvatier, Allan Dafoe, Baobao Zhang, and Owain Evans. 2018.
\newblock When will ai exceed human performance? evidence from ai experts.
\newblock \emph{Journal of Artificial Intelligence Research}, 62:729--754.

\bibitem[{Gr{\"o}ger(2021)}]{groger2021there}
Christoph Gr{\"o}ger. 2021.
\newblock There is no ai without data.
\newblock \emph{Communications of the ACM}, 64(11):98--108.

\bibitem[{Huang et~al.(2015)Huang, Schuehle, Porter, and
  Youtie}]{huang2015systematic}
Ying Huang, Jannik Schuehle, Alan~L Porter, and Jan Youtie. 2015.
\newblock A systematic method to create search strategies for emerging
  technologies based on the web of science: illustrated for ‘big data’.
\newblock \emph{Scientometrics}, 105:2005--2022.

\bibitem[{Ji et~al.(2023)Ji, Lee, Frieske, Yu, Su, Xu, Ishii, Bang, Madotto,
  and Fung}]{ji2023survey}
Ziwei Ji, Nayeon Lee, Rita Frieske, Tiezheng Yu, Dan Su, Yan Xu, Etsuko Ishii,
  Ye~Jin Bang, Andrea Madotto, and Pascale Fung. 2023.
\newblock Survey of hallucination in natural language generation.
\newblock \emph{ACM Computing Surveys}, 55(12):1--38.

\bibitem[{Kim et~al.(2023)Kim, Guo, Yu, and Li}]{kim2023can}
Yuheun Kim, Lu~Guo, Bei Yu, and Yingya Li. 2023.
\newblock Can chatgpt understand causal language in science claims?
\newblock In \emph{Proceedings of the 13th Workshop on Computational Approaches
  to Subjectivity, Sentiment, \& Social Media Analysis}, pages 379--389.

\bibitem[{Krafft et~al.(2020)Krafft, Young, Katell, Huang, and
  Bugingo}]{krafft2020defining}
PM~Krafft, Meg Young, Michael Katell, Karen Huang, and Ghislain Bugingo. 2020.
\newblock Defining ai in policy versus practice.
\newblock In \emph{Proceedings of the AAAI/ACM Conference on AI, Ethics, and
  Society}, pages 72--78.

\bibitem[{Kurzweil(1985)}]{kurzweil1985artificial}
Raymond Kurzweil. 1985.
\newblock What is artificial intelligence anyway? as the techniques of
  computing grow more sophisticated, machines are beginning to appear
  intelligent—but can they actually think?
\newblock \emph{American Scientist}, 73(3):258--264.

\bibitem[{Mart{\i}nez-Plumed et~al.(2018)Mart{\i}nez-Plumed, Loe, Flach,
  O~hEigeartaigh, Vold, and Hern{\'a}ndez-Orallo}]{martinez2018facets}
Fernando Mart{\i}nez-Plumed, Bao~Sheng Loe, Peter Flach, Se{\'a}n
  O~hEigeartaigh, Karina Vold, and Jos{\'e} Hern{\'a}ndez-Orallo. 2018.
\newblock The facets of artificial intelligence: A framework to track the
  evolution of ai.
\newblock In \emph{International Joint Conferences on Artificial Intelligence},
  pages 5180--5187.

\bibitem[{Miyazaki and Sato(2018)}]{miyazaki2018analyses}
Kumiko Miyazaki and Ryusuke Sato. 2018.
\newblock Analyses of the technological accumulation over the 2 nd and the 3 rd
  ai boom and the issues related to ai adoption by firms.
\newblock In \emph{2018 Portland International Conference on Management of
  Engineering and Technology (PICMET)}, pages 1--7. IEEE.

\bibitem[{Mogoutov and Kahane(2007)}]{mogoutov2007data}
Andrei Mogoutov and Bernard Kahane. 2007.
\newblock Data search strategy for science and technology emergence: A scalable
  and evolutionary query for nanotechnology tracking.
\newblock \emph{Research Policy}, 36(6):893--903.

\bibitem[{Mustafa et~al.(2021)Mustafa, Usman, Yu, Afzal, Sulaiman, and
  Shahid}]{mustafa2021multi}
Ghulam Mustafa, Muhammad Usman, Lisu Yu, Muhammad~Tanvir Afzal, Muhammad
  Sulaiman, and Abdul Shahid. 2021.
\newblock Multi-label classification of research articles using word2vec and
  identification of similarity threshold.
\newblock \emph{Scientific Reports}, 11(1):21900.

\bibitem[{Niu et~al.(2016)Niu, Tang, Xu, Zhou, and Song}]{niu2016global}
Jiqiang Niu, Wenwu Tang, Feng Xu, Xiaoyan Zhou, and Yanan Song. 2016.
\newblock Global research on artificial intelligence from 1990--2014:
  Spatially-explicit bibliometric analysis.
\newblock \emph{ISPRS International Journal of Geo-Information}, 5(5):66.

\bibitem[{OpenAI(2023)}]{openai2023gpt4}
OpenAI. 2023.
\newblock \href {http://arxiv.org/abs/2303.08774} {Gpt-4 technical report}.

\bibitem[{Priem et~al.(2022)Priem, Piwowar, and Orr}]{priem2022openalex}
Jason Priem, Heather Piwowar, and Richard Orr. 2022.
\newblock Openalex: A fully-open index of scholarly works, authors, venues,
  institutions, and concepts.
\newblock \emph{arXiv preprint arXiv:2205.01833}.

\bibitem[{Russell(2010)}]{russell2010artificial}
Stuart~J Russell. 2010.
\newblock \emph{Artificial intelligence a modern approach}.
\newblock Pearson Education, Inc.

\bibitem[{Sachini et~al.(2022)Sachini, Sioumalas-Christodoulou, Christopoulos,
  and Karampekios}]{sachini2022ai}
Evi Sachini, Konstantinos Sioumalas-Christodoulou, Stefanos Christopoulos, and
  Nikolaos Karampekios. 2022.
\newblock Ai for ai: Using ai methods for classifying ai science documents.
\newblock \emph{Quantitative Science Studies}, pages 1--14.

\bibitem[{Schank(1987)}]{schank1987ai}
Roger~C Schank. 1987.
\newblock What is ai, anyway?
\newblock \emph{AI magazine}, 8(4):59--59.

\bibitem[{Shen et~al.(2018)Shen, Ma, and Wang}]{shen-etal-2018-web}
Zhihong Shen, Hao Ma, and Kuansan Wang. 2018.
\newblock \href {https://doi.org/10.18653/v1/P18-4015} {A web-scale system for
  scientific knowledge exploration}.
\newblock In \emph{Proceedings of {ACL} 2018, System Demonstrations}, pages
  87--92, Melbourne, Australia. Association for Computational Linguistics.

\bibitem[{Shukla et~al.(2019)Shukla, Janmaijaya, Abraham, and
  Muhuri}]{shukla2019engineering}
Amit~K Shukla, Manvendra Janmaijaya, Ajith Abraham, and Pranab~K Muhuri. 2019.
\newblock Engineering applications of artificial intelligence: A bibliometric
  analysis of 30 years (1988--2018).
\newblock \emph{Engineering Applications of Artificial Intelligence},
  85:517--532.

\bibitem[{Sinha et~al.(2015)Sinha, Shen, Song, Ma, Eide, Hsu, and
  Wang}]{sinha2015}
Arnab Sinha, Zhihong Shen, Yang Song, Hao Ma, Darrin Eide, Bo-June~(Paul) Hsu,
  and Kuansan Wang. 2015.
\newblock \href {https://doi.org/10.1145/2740908.2742839} {An overview of
  microsoft academic service (mas) and applications}.
\newblock In \emph{Proceedings of the 24th International Conference on World
  Wide Web}, WWW '15 Companion, page 243–246, New York, NY, USA. Association
  for Computing Machinery.

\bibitem[{Suominen and Newman(2017)}]{suominen2017exploring}
Arho Suominen and Nils~C Newman. 2017.
\newblock Exploring the fundamental conceptual units of technical emergence.
\newblock In \emph{2017 Portland International Conference on Management of
  Engineering and Technology (PICMET)}, pages 1--5. IEEE.

\bibitem[{Sweeney()}]{sweeney2003s}
Latanya Sweeney.
\newblock That's ai?: a history and critique of the field.

\bibitem[{Toney and Dunham(2022)}]{toney2022multi}
Autumn Toney and James Dunham. 2022.
\newblock Multi-label classification of scientific research documents across
  domains and languages.
\newblock In \emph{Proceedings of the Third Workshop on Scholarly Document
  Processing}, pages 105--114.

\bibitem[{Wang et~al.(2021)Wang, Liu, Xu, Zhu, and
  Zeng}]{wang-etal-2021-want-reduce}
Shuohang Wang, Yang Liu, Yichong Xu, Chenguang Zhu, and Michael Zeng. 2021.
\newblock \href {https://doi.org/10.18653/v1/2021.findings-emnlp.354} {Want to
  reduce labeling cost? {GPT}-3 can help}.
\newblock In \emph{Findings of the Association for Computational Linguistics:
  EMNLP 2021}, pages 4195--4205, Punta Cana, Dominican Republic. Association
  for Computational Linguistics.

\bibitem[{West and Allen(2018)}]{west2018artificial}
Darrell~M West and John~R Allen. 2018.
\newblock How artificial intelligence is transforming the world.
\newblock \emph{Report. April}, 24:2018.

\bibitem[{Woolridge(2022)}]{woolridge}
Micheal Woolridge. 2022.
\newblock \emph{A brief history of artificial intelligence: what it is, where
  we are, and where we are going}.
\newblock Flatiron Books.

\bibitem[{Zhang et~al.(2021)Zhang, Mishra, Brynjolfsson, Etchemendy, Ganguli,
  Grosz, Lyons, Manyika, Niebles, Sellitto et~al.}]{zhang2021ai}
Daniel Zhang, Saurabh Mishra, Erik Brynjolfsson, John Etchemendy, Deep Ganguli,
  Barbara Grosz, Terah Lyons, James Manyika, Juan~Carlos Niebles, Michael
  Sellitto, et~al. 2021.
\newblock The ai index 2021 annual report.
\newblock \emph{arXiv preprint arXiv:2103.06312}.

\end{thebibliography}

\end{document}